\documentclass[11pt, a4paper, logo, twocolumn]{brown}
\usepackage[toc,page]{appendix}

\usepackage{times}

\usepackage{multirow}
\usepackage{multicol}
\usepackage{tablefootnote} %
\usepackage{caption}
\usepackage{subcaption}
\usepackage{graphicx}
\usepackage{float}
\usepackage{amssymb}
\usepackage{amsmath}

\usepackage{lipsum} 
\usepackage{tabularx}
\usepackage{adjustbox}
\newcommand{\autosizeTable}[1]{%
    \begin{adjustbox}{max width=\linewidth}
        #1
    \end{adjustbox}
}

\usepackage{sidecap}
\usepackage{hyphenat}
\usepackage{makecell}
\usepackage{booktabs}
\usepackage{xspace}

\newcommand{\shortname}{UniTac\xspace}
\newcommand{\networkname}{UniTac-Net\xspace}

\definecolor{LightGray}{gray}{0.7}
\definecolor{gold}{RGB}{255, 199, 44}
\colorlet{gold}{gold!30!white}
\newcolumntype{g}{>{\color{LightGray}}r}

\usepackage{pifont}%

\begin{document}
\title{\shortname: Whole-Robot Touch Sensing Without Tactile Sensors}
\runningtitle{\shortname: Whole-Robot Touch Sensing Without Tactile Sensors}

\author[1*]{Wanjia Fu}
\author[1*]{Hongyu Li}
\author[1]{Ivy X. He}
\author[1]{Stefanie Tellex}
\author[1]{Srinath Sridhar}
\affil[1]{Brown University} 
\affil[*]{Equal contribution} 
\correspondingauthor{Hongyu Li (\href{mailto:hli230@cs.brown.edu}{hli230@cs.brown.edu})}

\begin{abstract}
Robots can better interact with humans and unstructured environments through touch sensing.
However, most commercial robots are not equipped with tactile skins, making it challenging to achieve even basic touch-sensing functions, such as contact localization.
We present UniTac, a data-driven whole-body touch-sensing approach that uses only proprioceptive joint sensors and does not require the installation of additional sensors.  
Our approach enables a robot equipped solely with joint sensors to localize contacts.
Our goal is to democratize touch sensing and provide an off-the-shelf tool for HRI researchers to provide their robots with touch-sensing capabilities.
We validate our approach on two platforms: the Franka robot arm and the Spot quadruped.  
On Franka, we can localize contact to within 8.0 centimeters, and on Spot, we can localize to within 7.2 centimeters at around 2,000 Hz on an RTX 3090 GPU without adding any additional sensors to the robot.
Project website: \url{https://ivl.cs.brown.edu/research/unitac}.
\end{abstract}

\twocolumn[{%
\renewcommand\twocolumn[1][]{#1}%
\maketitle
\begin{center}
    \centering
    \captionsetup{type=figure}
    \includegraphics[width=\textwidth]{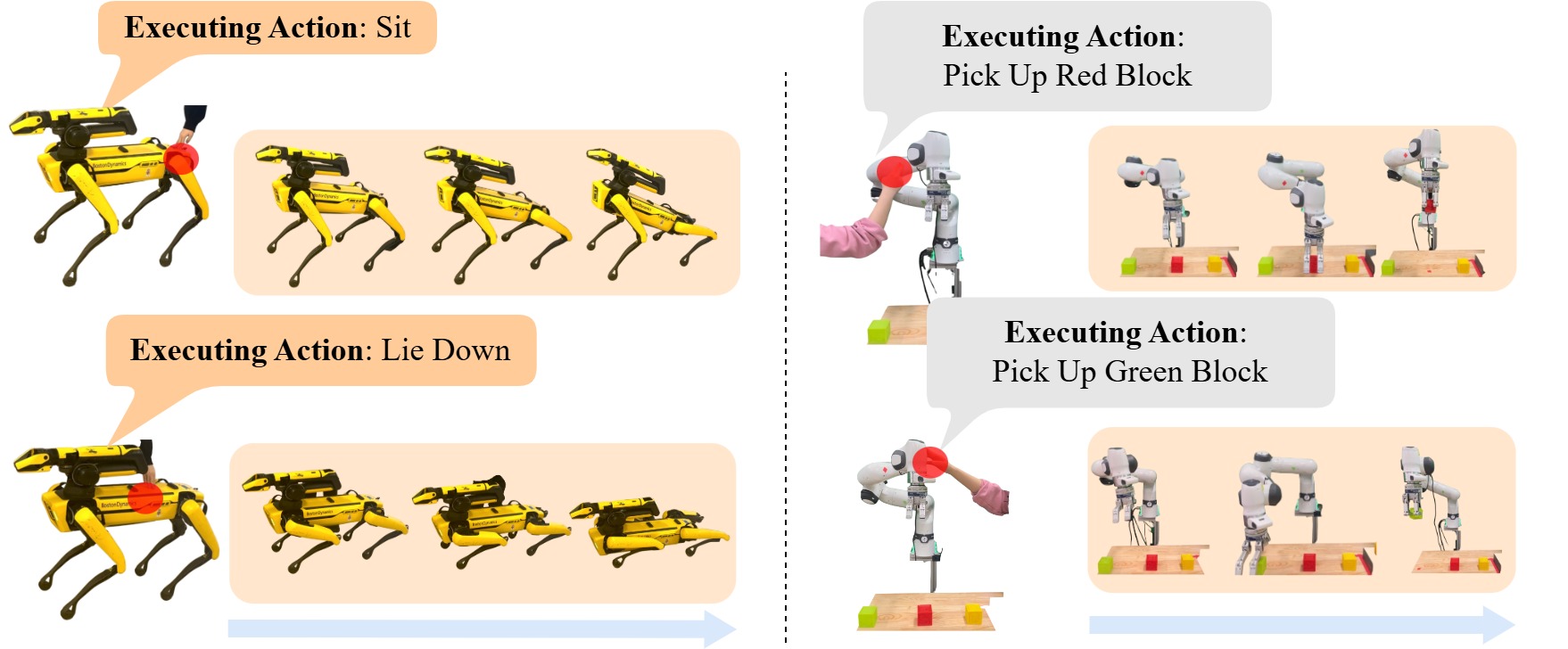}
    \caption{\textbf{Interactions achieved using \shortname}.
    \shortname achieves whole-robot touch sensing without using any tactile sensors and empowers applications such as patting the quadruped for canine-inspired responses or touch-based instructions for manipulation.
    Our method can be applied to robots with different embodiment types, including quadrupeds and arms.
    }
    \label{fig:teaser}
\end{center}
}]

\section{Introduction}

Commercial robots are becoming increasingly capable.
We now have bipedal/quadrupedal robots that can walk or run in challenging environments~\cite{valsecchi_quadrupedal_2020, song_learning_2024, agarwal_legged_2022}, and robot arms that assemble products with precision~\cite{haddadin_franka_2022}.
Despite the impressive capabilities of these robots, they lack a critical aspect of animal behavior: physical interaction through touch (Fig.~\ref{fig:teaser}).
Consider how a simple pat can convey trust or instruction when interacting with a person or an animal~\cite{gallace2010science}.
A similar level of nuanced touch-based communication is currently out of reach for most robots.
This limitation is largely due to the absence or difficulty in endowing robots with touch-sensing capabilities.
Although hardware and software advances for touch sensing have been made,  
the practical challenges of integrating tactile sensors into robots has prevented widespread use~\cite{dahiya_tactile_2010, yu_recent_2024}.

Touch sensing is essential for a variety of tasks, including recovery from unintended collisions~\cite{iwate_human-humanoid_2000, liang_contact_2021}, classification of types of human touch~\cite{koo_online_2008}, improving visual state estimation~\cite{li_vihope_2023, li_vita-zero_2025, li_v-hop_2025}, provision of social support for the elderly~\cite{block_six_2021}, and aid in minimally invasive surgery~\cite{tiwana_review_2012}. 
These interactions have been supported by installing dedicated tactile sensors, including sensors on robot hands~\cite{lambeta_digit_2020, lambeta_digitizing_2024, li_hypertaxel_2024}, and full-body tactile skins, either rigid~\cite{iwate_human-humanoid_2000, frigola_human-robot_2006, koo_online_2008, bhirangi_anyskin_2024} or soft~\cite{iwata_human-robot-contact-state_2005, ishiguro_robovie_2001, kanda_development_2002, thrun_haptic_2007, mitsunaga_robovie-iv_2006, tajika_reducing_2008, minato_evaluating_2006, zhong_dynamics-oriented_2025}.
Despite their usefulness, installing these tactile sensors is error-prone and cumbersome.
Rigid sensors tend to compromise the robot’s dexterity and mechanical flexibility, while soft sensors are prone to damage and produce errors due to self-contact at joints~\cite{argall_survey_2010}. 
Moreover, the integration of tactile sensors involves high costs and complex considerations such as calibration, power supply, wiring, and communication infrastructure~\cite{zhao_-situ_2023, dahiya_tactile_2010}.
Alternative methods that enable robots to feel touch without dedicated tactile sensors would address these issues.  

In this paper, we propose \textbf{\shortname}, a unified method to enable \textbf{whole-robot touch sensing capabilities across different robots without tactile sensors}.
Our approach is applicable to various robot platforms and leverages data from \emph{only existing sensors}. 
Specifically, we use torque and position data from joint sensors, which are readily available on most commercial robots.
Different contact patterns generate distinguishable proprioceptive feedback, which can be used to infer the location of contact.
Unlike model-based approaches~\cite{iskandar_intrinsic_2024, manuelli_localizing_2016}, which depend on physical models and demand extensive expert tuning for each robot platform, our technique is entirely data-driven. 
While prior data-driven methods~\cite{zwiener_contact_2018, liang_contact_2021} depend on simulated data - necessitating the construction of bespoke simulations for each robot - we train a neural network on real-world joint sensor data to directly predict contact location in real time, thereby eliminating simulation designs and the sim-to-real gap.
Notably, our efficient data collection process requires as little as 2.5 hours (for Spot), yet it is sufficient for robust real-world whole-robot touch sensing.

To validate our approach, we collected datasets from both a robot arm  (Franka Research 3 or FR3) and a quadruped platform (Boston Dynamic Spot), applying touch at sampled points (Fig.~\ref{fig:spot_sample}) across the robots' surfaces.
We introduced random perturbations in joint positions between data collections to ensure robustness.
Using only the joint sensors, \shortname is able to localize touches within an average of 7.2 centimeters on Spot and 8.0 centimeters on Franka at a runtime speed of 2,000 Hz.  
We further conducted experiments on another Spot robot, which was not used during data collection.
Results confirm the generalizability of \shortname on different robot instances with the same type.

We demonstrate potential applications of \shortname in physical Human-Robot Interaction (pHRI)~\cite{farajtabar_path_2024}.
For instance, with live predictions from \shortname, a quadruped robot (Spot) can react to human touch in real time, while a Franka robotic arm can perform manipulation tasks based on the location of human contact (Fig.~\ref{fig:teaser}).
\noindent\textbf{Limitations \& Future Work:} Although our work is currently limited to single contact localization, we believe that it unlocks a new research direction and robot interaction applications without the need for cumbersome tactile sensors.

In summary, our contributions are as follows:

\begin{itemize} 
\item We present a data-driven model, \shortname, that leverages built-in joint torque sensors to achieve live whole-body touch sensing across
various robot platforms, eliminating the need for dedicated tactile sensors.
\item \shortname demonstrates generalizability across multiple robot instances with the same type, allowing a wider community to use it as an off-the-shelf interface directly.
\item We demonstrate potential applications in touch-based human-robot interaction, including scenarios such as bio-inspired quadruped choreography. 
\end{itemize}

\section{Related Works}

Prior works have studied the problem of using proprioceptive feedback for touching sensing.
These works localize the contact on the robot body and can be categorized into two main directions: model-driven and data-driven.

\subsection{Model-Driven Approaches}
Model-driven methods leverage explicit physical models to interpret sensor data for contact localization.
\citet{manuelli_localizing_2016} introduced the contact particle filter, which formulates contact localization as a convex quadratic optimization problem integrated with particle filtering and demonstrated on a simulated humanoid.
\citet{iskandar_intrinsic_2024} derived momentum-based equations from redundant joint force-torque sensor data, achieving high-resolution touch sensing without additional tactile hardware.
Although these approaches eliminate the need for extensive data collection, they demand significant expert knowledge to develop and fine-tune models for each specific robot platform and configuration, thereby limiting their adaptability.  
Furthermore, if the robot's payload changes, the model and model parameters will need to be updated by an expert, in contrast to our approach, which merely requires collecting an updated dataset.

\subsection{Data-Driven Approaches}
Data-driven methods offer an alternative by learning contact localization directly from sensor data.
Seminal works~\cite{zwiener_contact_2018, liang_contact_2021} have demonstrated this approach by simulating proprioceptive feedback.
\citet{zwiener_contact_2018} employed random forests and multi-layer perceptrons (MLP) to classify contacts at pre-selected points on the Jaco arm, while \citet{liang_contact_2021} trained a neural network to localize contact on the Franka arm by projecting contact points onto mesh surfaces.
Both studies follow the sim-to-real paradigm, in which they carefully design a simulation, train the model using simulated data, and deploy it on a single robot arm in the real world.
In contrast, our work collects data exclusively from real-world platforms, thereby closing the sim-to-real gap and eliminating the need for customizing simulation environments.  
Moreover, our approach is applicable to various embodiments, including both robot arms and quadrupeds, and provides a straightforward interface for HRI researchers in developing downstream capabilities.

\section{Method}
\label{methodology}

\begin{figure}
    \centering
        \includegraphics[width=\linewidth]{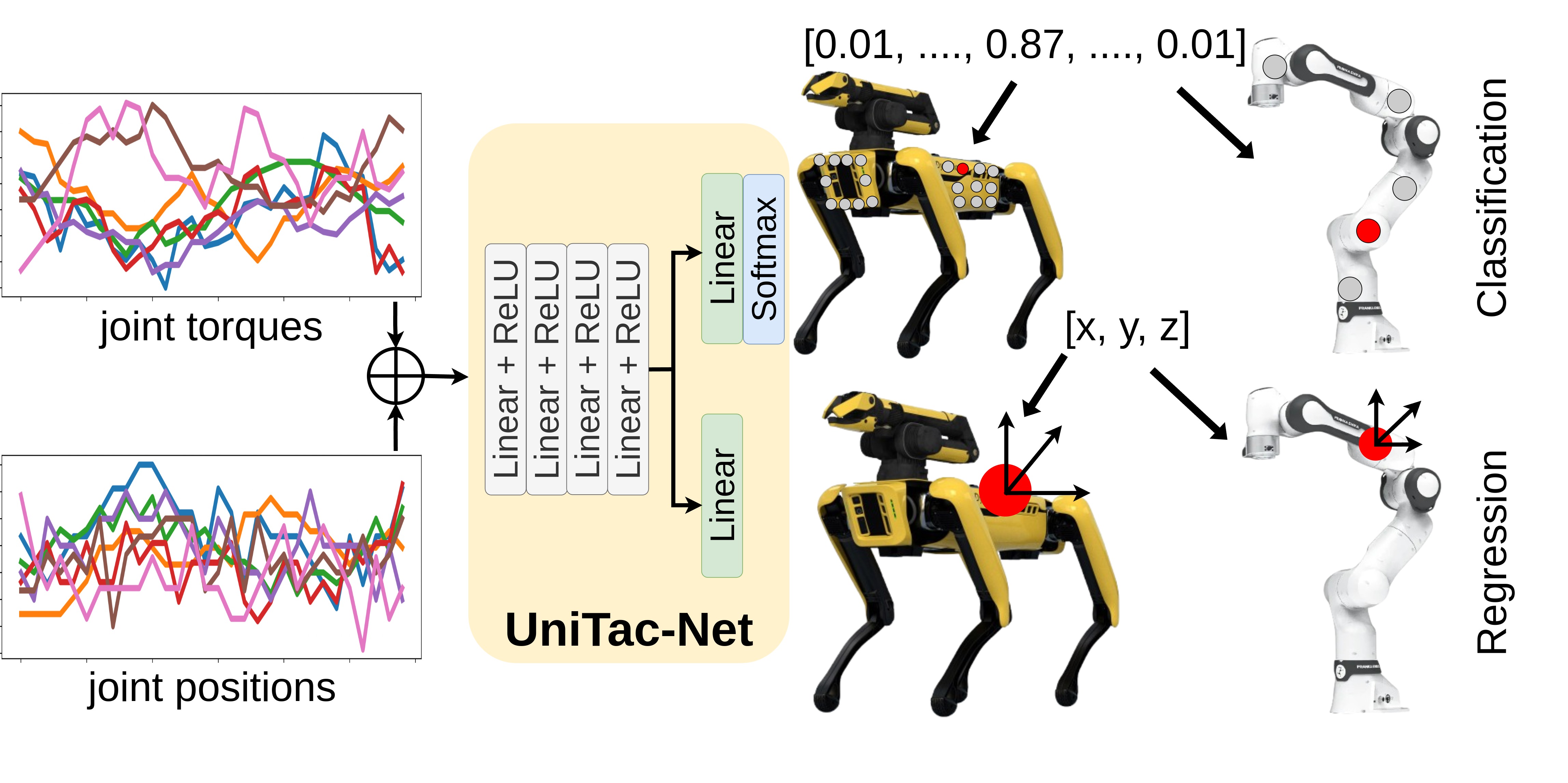}
    \caption{
    \textbf{Design of \networkname.}
    \networkname is a four-layer MLP with either a regression or a classification output head.
    It takes proprioceptive feedback (joint torques and positions) as input and predicts the contact (if any).
    } 
    \label{fig:network}
\end{figure}

Our goal is to develop a method for localizing touch on the robot's surface using only proprioceptive feedback. 
We first randomly sample a preset number of $n$ points on the surface of the robot mesh and define them as the ground truth contact locations.
We collect joint data during contact at each point multiple times by varying joint configurations, and construct a dataset $\mathcal{D} = \{ d_1, d_2, \dots, d_k \}$ with $k$ samples.
A detailed process for contact collection will be explained in the next section.
Each data tuple is $d_i = (p_i,q_i,\tau_i)$, where $p \in \mathbb{R}^3$ is the ground truth contact location. 
$q \in \mathbb{R}^{DoF}$ and $\tau \in \mathbb{R}^{DoF}$ are the joint positions and torques, respectively.
We build a contact localization model that maps the proprioceptive signal - joint positions and torques ($q$ and $\tau$) - to the contact coordinate ($p$), defined in the robot frame, using a neural network, namely \textbf{\networkname} (Fig.~\ref{fig:network}). 
Contact localization can be treated as either a regression or a classification problem~\cite{molchanov_contact_2016, liang_contact_2021}, which differ in the output head of the neural network.

\subsection{Regression}
Our regression approach directly predicts the continuous 3D coordinates of the contact point.
For the architecture, we use a four-layer MLP with layer sizes of 64, 128, 256, and 128, respectively, and a dropout rate of 0.3 after each layer.
ReLU activations are used in the hidden layers, and the final layer has a size of three ($x,y,z$ coordinates) and no activation function.
A ``no-contact'' state is represented by the point (0, 0, 0). 
The model is trained using the mean squared error loss between ground truth $p$ and predicted $\hat{p}$ as the loss.
Compared to the classification model, the regression method avoids the inherent discretization error, potentially providing more precise localization. 
However, it may be more sensitive to noise in sensor readings, necessitating careful regularization and filtering.

Our empirical results suggest that the regression model performs better than the classification model. 
Therefore, we resort to all the models used in the future sections as the regression model.

\subsection{Classification}
When we treat it as a classification problem, we predict one of the pre-defined $n$ points (as classes) with an additional ``no-contact'' class. 
The ground truth labels are obtained by mapping each ground truth contact location $p$ into a one-hot encoding $c$ with size of $(n+1)$.
The classification model shares the same overall structure as the regression model up to the final layer. In this case, the final output layer uses a softmax activation to produce a probability distribution over the classes.
We compute the cross-entropy loss between the ground truth class index $c$ and the one-hot encoding for the predicted class index $\hat{c}$.

The classification method is robust against minor variations in joint signals because it maps inputs to a limited set of well-defined contact locations. 
However, it discretizes the contact space, which may limit precision if the contact point falls between the sampled points. 
The choice of the number of classes $(n+1)$ plays a crucial role here: increasing it can improve spatial resolution but may also make the model more complex and prone to overfitting.

\subsection{Training}
We preprocess the data by normalizing the joint positions $q$ and torques $\tau$ to the range of -1 to 1.
This normalization, performed for each dimension across the entire dataset, helps standardize the input and speeds up the convergence of the training process.
Both models are trained using the Adam optimizer~\cite{kingma_adam_2017} with a fixed learning rate of $2.5\times10^{-3}$.
Training is conducted over 30 epochs with a batch size of 256.

\subsection{Live Detection}
\networkname could run at the speed of 2,000 Hz on an NVIDIA RTX 3090 GPU.
However, such a high frequency could lead to unstable predictions.
Therefore, we apply an exponential moving average (EMA) filter on model predictions with a smoothing factor of 0.1 and a sliding window length of 40. 
This EMA filter is used to filter noise in the output and improve temporal stability.

\begin{figure}
    \centering
    \begin{subfigure}{0.29\textwidth}
        \centering
        \includegraphics[width=\linewidth]{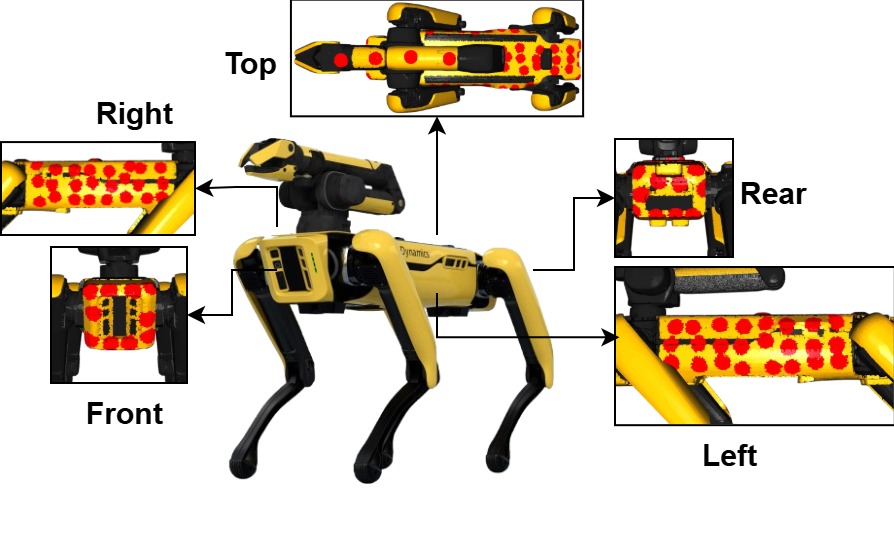}
    \end{subfigure}
    \begin{subfigure}{0.18\textwidth}
        \centering
        \includegraphics[width=\linewidth]{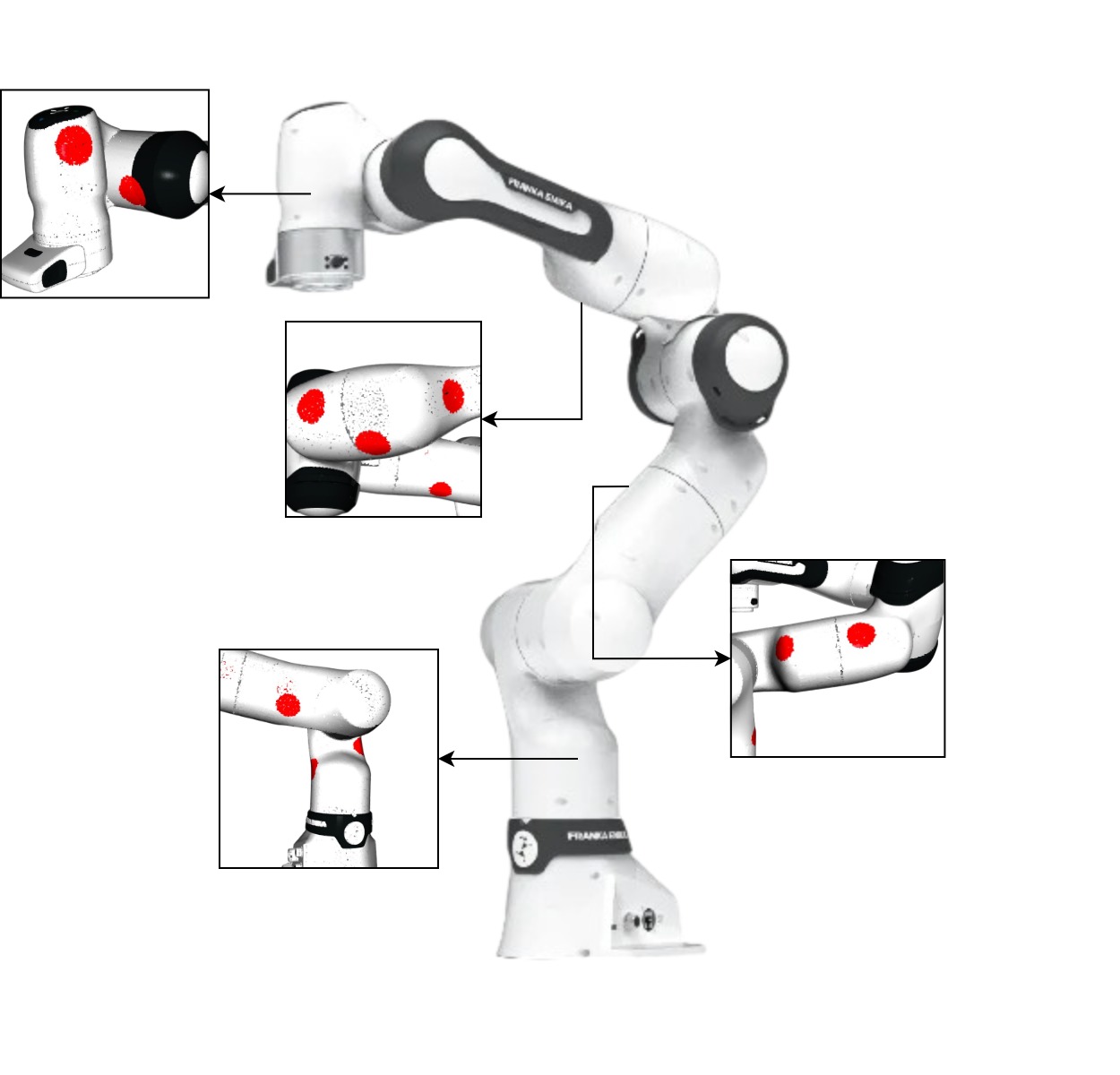}
    \end{subfigure}
    \caption{
    \textbf{Sampled contact points for data collection.}
    We sample 104 points on Spot (left) and 10 points on Franka (right).
    Dense sampling on Spot covers the whole robot except for the legs, while the sparser sampling on Franka covers each link.
    }
    \label{fig:spot_sample}
\end{figure}

\begin{figure*}
    \centering
     \includegraphics[width=\textwidth]{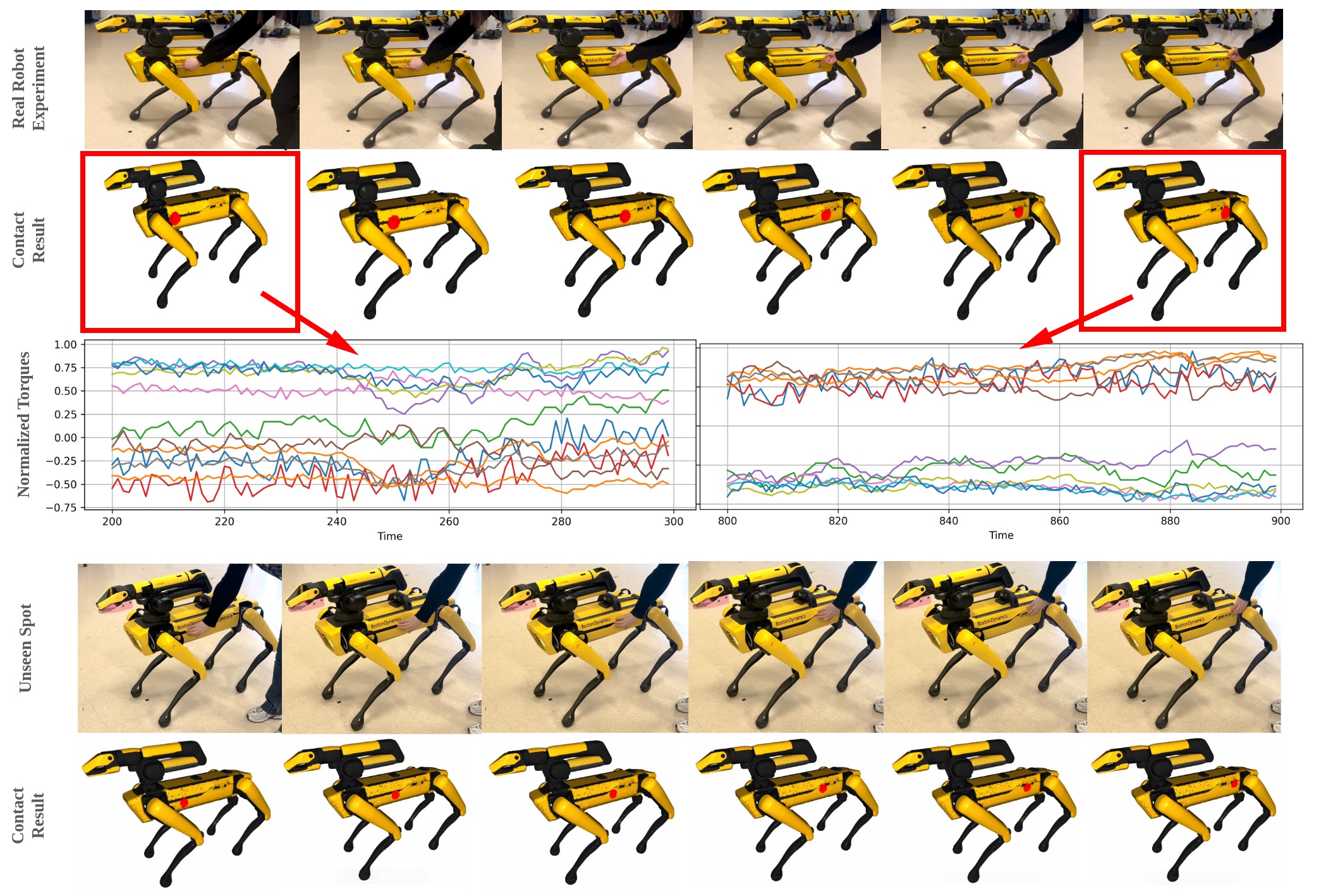}
    \caption{
    \textbf{Live contact localization on Spots.}
    Top row: A human applies touch to the robot.
    Middle row: The system localizes the contact point on the robot’s mesh.
    Bottom row: Normalized joint torque changes are displayed (different colors indicate distinct joint sensors).
    The last two rows show that similar localization performance is achieved on \emph{an unseen Spot} not used during training. 
    }
    \label{fig:torques_spot}
\end{figure*}

\begin{figure*}
    \centering
     \includegraphics[width=\linewidth]{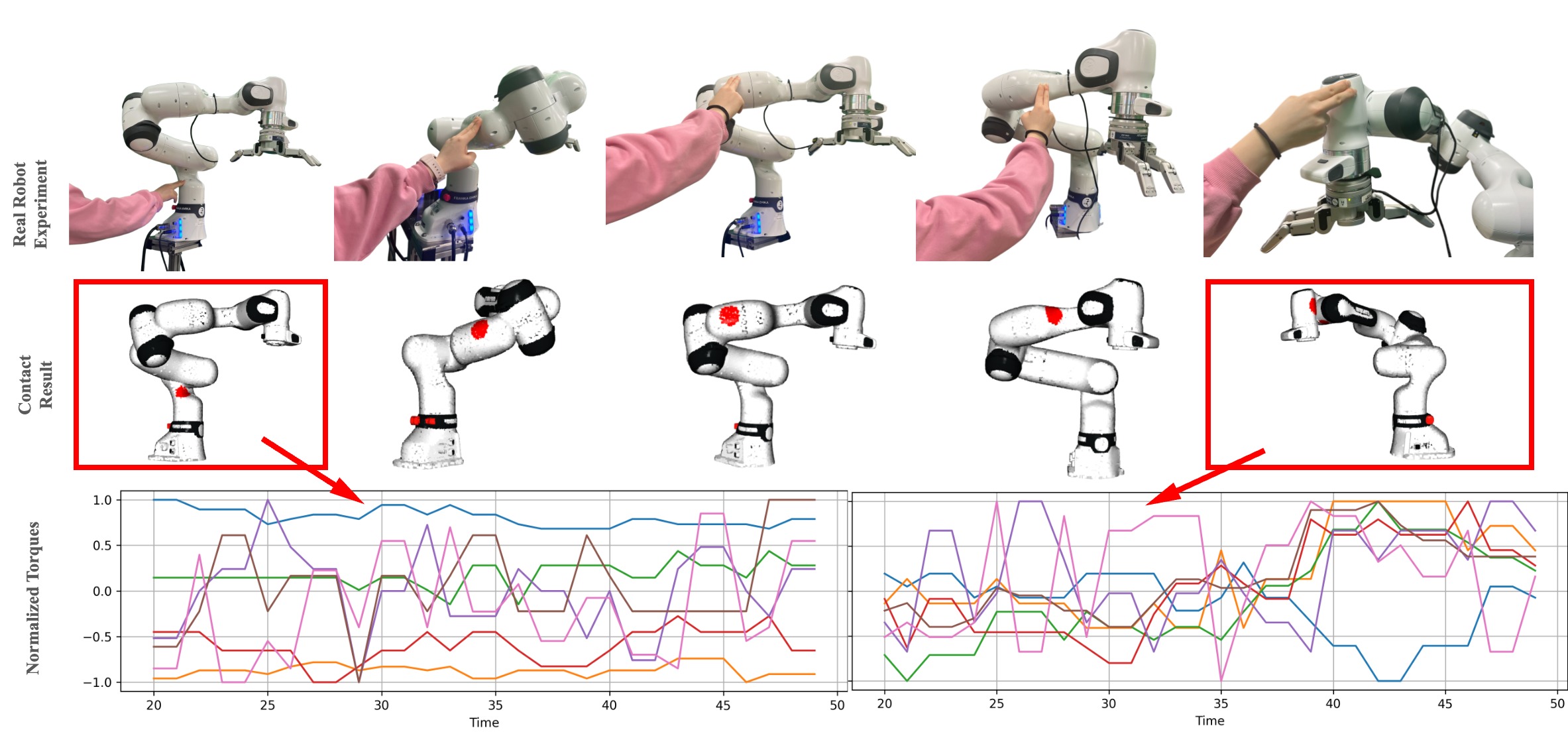}
    \caption{
    \textbf{Live contact localization on FR3 robot arm.}
    Similar to the Spot figure above, we show the actual human touch, localization result, and torque value change.
    The torque plots show the unique signature look for each touch location, which we utilized to localize the contact.
    More results are provided in our supplementary video.
    }
    \label{fig:torques_franka}
\end{figure*}

\section{Experiments}
We validate the effectiveness of \shortname on two platforms with distinct morphologies: the Spot quadruped from Boston Dynamics and the Franka Research 3 robotic arm from Franka Robotics.
In this section, we first introduce our data collection process.
Based on the collected dataset, we conduct experiments to evaluate the performance of our contact localization model.
In the end, we demonstrate pHRI applications based on real-time deployment of \shortname.

\subsection{Data Collection}
\label{exp:data}

\subsubsection{Spot}
Spot (with robot arm) has a total of 19 degrees of freedom (DoF): 12 joints for the legs (3 per leg), 6 for the arm, and 1 for the gripper. 
In our experiments, we focused on detecting where contact occurs on the robot’s body and arm while ignoring the legs for safety reasons.
We sampled 104 contact points on the robot. 
These points are distributed as follows: 26 points each on the left, right, and top surfaces; 12 points on the back; 10 on the front; and 4 on the arm, as shown in Fig.~\ref{fig:spot_sample} (left). 
The robot is kept stationary for each sample.
To capture different configurations, we intentionally randomize the joint positions by teleoperating the robot into various postures (including different standing heights).
Joint states for Spot are recorded using the Boston Dynamics Clients API at 60Hz. 

\subsubsection{Franka}

FR3 model has 7 DoF. 
Unlike the quadruped Spot, the FR3 is more challenging for contact localization because it has an open kinematic chain with a fixed base.
The robot arm joints are arranged sequentially, as opposed to the quadruped.
We sample 10 points on the surface of FR3, as shown in Fig.~\ref{fig:spot_sample} (right).
The arm is assumed to be equipped with a fixed end-effector that remains steady throughout the experiments. (Changing the end-effector might lead to worse performance, since our model was trained using this specific configuration.)
Similar to Spot, we perturb the joint positions after each collection.
Since FR3 has an open kinematic chain and a fixed base, it has significantly larger joint space.
To keep our dataset size plausible, we keep the end-effector roughly in the same area while changing the joints.
Joint torques and joint positions for FR3 are recorded using \texttt{libfranka} SDK at 30Hz.

Data is first collected when no contact is made with the robot, in which case the robot remains stationary. Meanwhile, the visualizer window shows the robot mesh to prepare the user for the following steps. 
The user will then collect each selected contact location, where the visualizer shows the robot mesh with a red marker on the ground truth touch location. 
For data augmentation purposes, when collecting each sample, the user may vary the direction and magnitude of the touch force they apply as long as the touch location is kept the same.
Spot data consists of joint state data during which contact is applied at 104 different touch locations throughout its whole body, collected over 50 sets of different joint configurations for 19 joints. 
In addition, data collected for FR3 consists of joint torque and joint position data during which contact is applied at 10 different touch locations across all arm links, collected over 25 sets of different joint configurations for 7 joints.
These take only 2.5 hours of data collection on Spot and 12 minutes on Franka. 

\subsection{Contact Localization}
\label{exp:local}

\subsubsection{Quantitative Results}
We use two metrics to evaluate the performance of our model: L2 norm and accuracy.
L2 norm is defined as the Euclidean distance $||p - \hat{p}||_2$ between the predicted position $\hat{p}$ and ground truth contact position $p$.
Accuracy (Acc) is calculated as the percentage of predictions whose Euclidean distance from the ground truth is within a threshold $\epsilon$:
\begin{equation}
    \text{Acc} = \frac{1}{N} \sum_{i=1}^{N} \mathbb{I}(\| p_i - \hat{p}_i \|_2 \le \epsilon ),
\end{equation}
where $N$ represents the number of samples.

\begin{figure}
    \centering
    \captionsetup{type=figure}
     \includegraphics[width=1.0\linewidth]{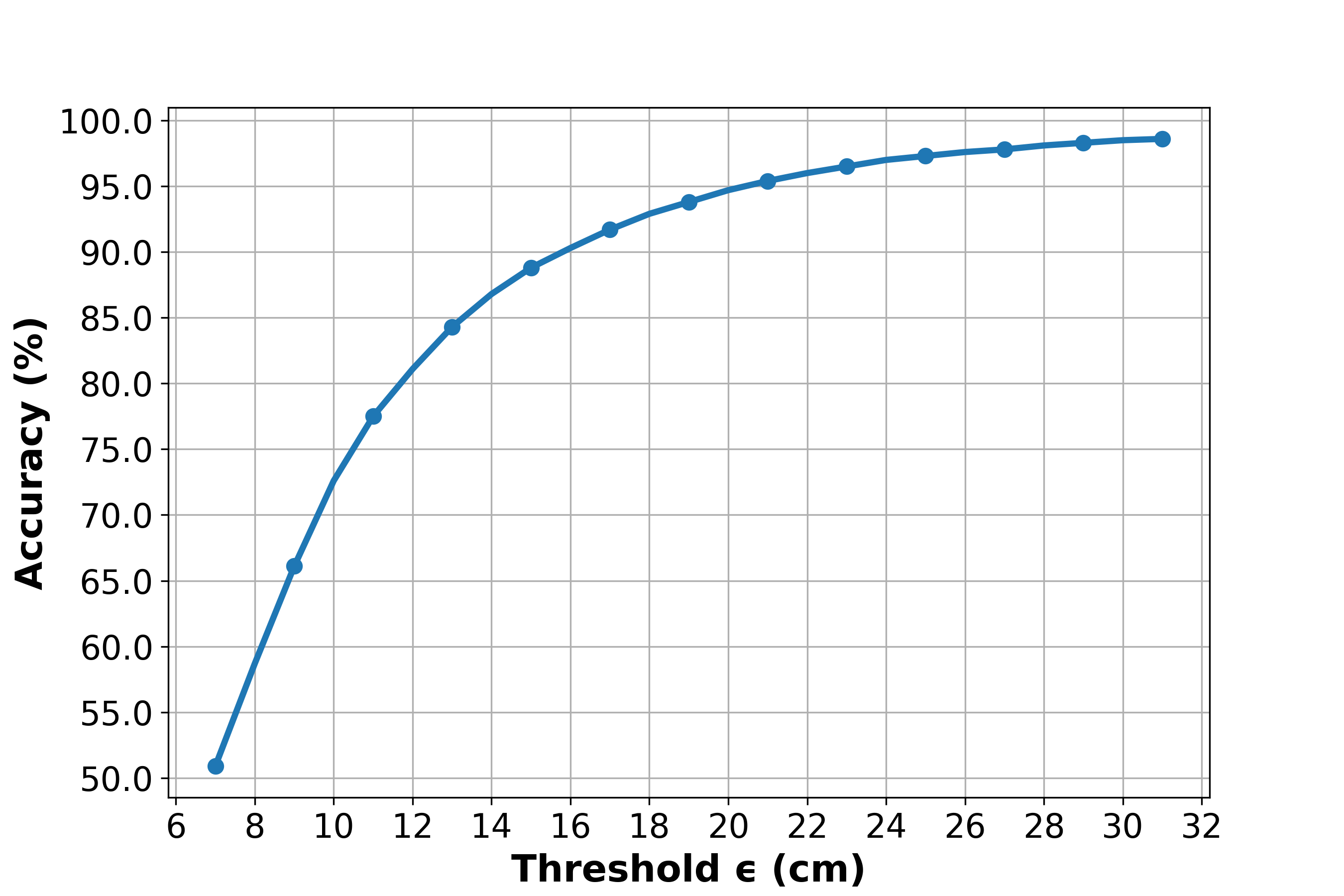}
    \caption{\textbf{Model accuracy with respect to threshold $\epsilon$}.
    We evaluate the accuracy of \networkname using various threshold values.
    }
    \label{fig:threshold}
\end{figure}

We compare our regression model with the classification model(Tab.~\ref{tab:comparison}).
When calculating the L2 norm for the classification model, the one-hot encodings are mapped back to the ground truth 3D coordinates of the contact positions. 
We partition the dataset into 80\% for training and 20\% for validation, and compare our regression model with the classification model, each evaluated alongside its corresponding k-nearest neighbors (KNN) baseline (with $k=3$, Tab.~\ref{tab:comparison}). 
\begin{table}[t]
\centering
\autosizeTable{
\begin{tabular}{l|r|r|r|r}
\hline
\multirow{2}{*}{Method} & \multicolumn{2}{c|}{Spot} & \multicolumn{2}{c}{Franka} \\
\cline{2-5}
 & Acc (\%) $\uparrow$ & L2 (cm) $\downarrow$ & Acc (\%) $\uparrow$ & L2 (cm) $\downarrow$ \\
\hline
KNN Classifier & 34.3 & \cellcolor{gold}24.1 & 55.8 & \cellcolor{gold}20.3 \\
KNN Regressor & 45.9 & \cellcolor{gold}19.4 & 43.3 & \cellcolor{gold}17.2 \\
Classification & 54.9 & \cellcolor{gold}13.7 & 53.7 & \cellcolor{gold}14.8 \\
Regression & \textbf{86.5} & \cellcolor{gold}\textbf{7.2} & \textbf{83.5} & \cellcolor{gold}\textbf{8.0} \\
\hline
\end{tabular}}
\caption{
\textbf{Comparison of model choices}.  
We compare the performance of our regression and classification models and KNN baselines.
}
\vspace{-2em}
\label{tab:comparison}
\end{table}
On Spot, the regression model achieves 86.5\% accuracy (at a threshold of $\epsilon=12$), with an average L2 error of 7.2 cm, compared to 54.9\% accuracy and a 13.7 cm L2 error for the classification model.
Similarly, on FR3, the regression approach attains 83.5\% accuracy with an 8.0 cm L2 error, while the classification method reaches only 53.7\% accuracy with a 14.8 cm L2 error.
Each method also outperforms its KNN baseline, demonstrating the effectiveness of our network design. Performance differences across robot platforms stem from differences in the number of sampled ground truth points and the amount of collected data.
These results demonstrate that the regression model outperforms the classification model by approximately 30 percentage points in accuracy and achieves a reduction of around 6.5 cm in L2 error.
This suggests that avoiding discretization leads to more precise contact localization, especially in scenarios with noisy sensor readings.
Fig.~\ref{fig:threshold} plots the model accuracy as a function of the threshold $\epsilon$. 

\subsubsection{Qualitative Results}
We demonstrate qualitative results of real-time contact localization on Spot (Fig.~\ref{fig:torques_spot}) and Franka (Fig.~\ref{fig:torques_franka}).
On the Spot, we slide our touch horizontally along the left side of the body and visualize the live contact localization prediction results.
Similarly, on the FR3 robot arm, we demonstrate by touching different links. 
The results demonstrate that our model could accurately localize rapidly changing contacts in real time. 

We also show that our real-time contact localization generalizes to different instances of the same robot model without additional retraining. As illustrated in the last two rows of Fig.~\ref{fig:torques_spot}, we slide our touch horizontally along the left side of an unseen Spot robot. The contact localization results exhibit similar accuracies with the results on a seen robot.

\begin{figure}
    \centering
        \includegraphics[width=\linewidth]{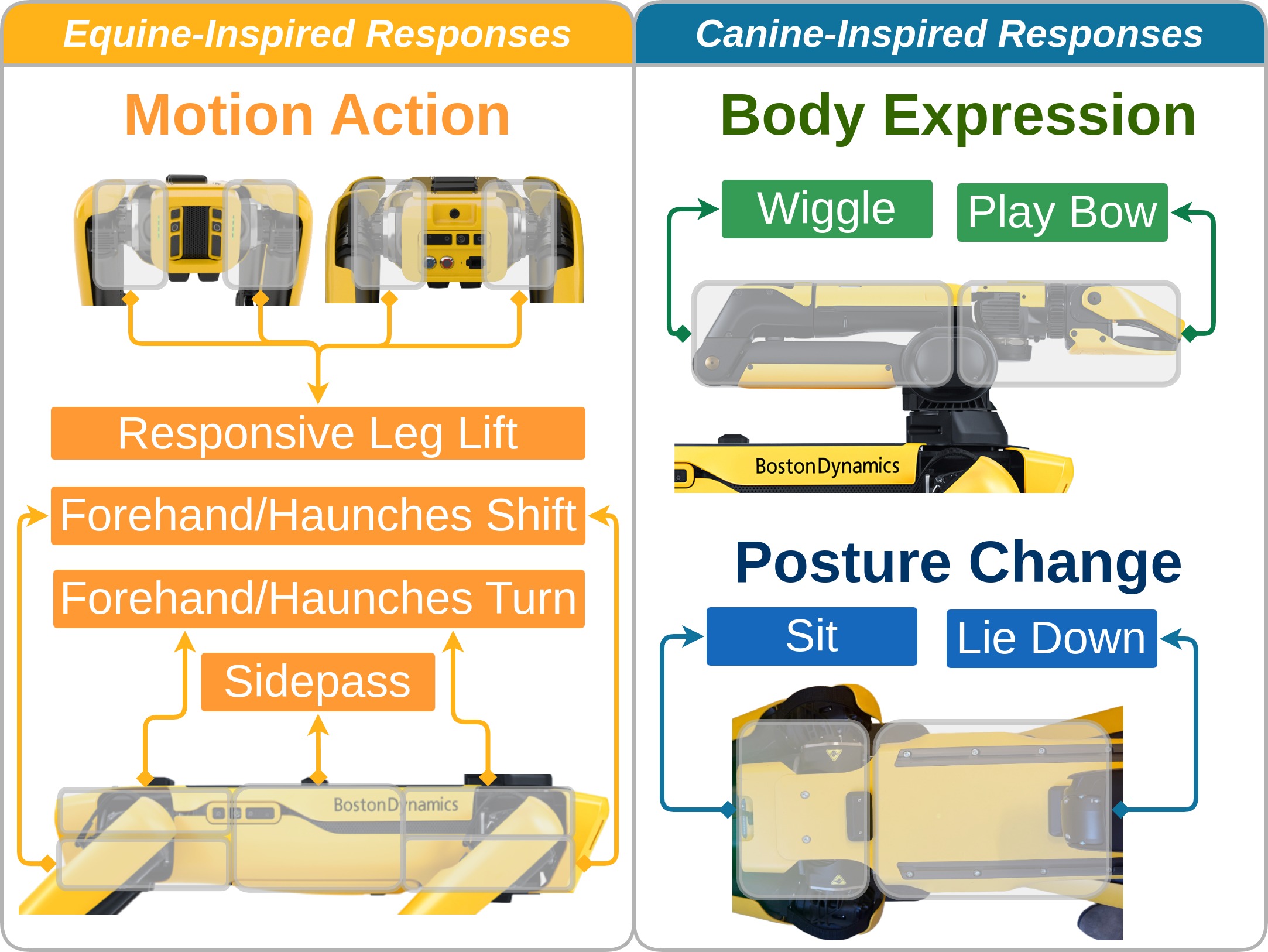}
    \caption{
    \textbf{Quadruped interactions design}.
     Touch-based interactions on Spot are categorized into three response types: motion actions (equine-inspired), posture changes, and body expressions (both canine-inspired)
    }
    \label{fig:spot_hri}
\end{figure}

\section{pHRI Applications}
\label{application:hri}
In the above sections, we presented how \shortname offers a pipeline to equip a quadruped or robot arm with whole-robot touch sensing capabilities. Tactile sensing enables more natural and intuitive pHRI by allowing robots to respond to physical touch. 
In this section, we further demonstrate potential applications enabled by such whole-robot touch-sensing abilities. 

\subsection{Quadruped Robot}
Humans have an innate understanding of animal behavior. 
Mimicking natural behaviors allows for more fluid interactions in scenarios like guiding, assisting, or responding to human intentions~\cite{zhan2023enable}.
To illustrate how a robot can behave and interact with humans in a way similar to animals with actual touch-sensing skins, we show quadruped choreography based on localized touch. 
This interaction design is inspired by ethologically relevant interactions observed in human-animal communication, particularly with social animals like dogs and horses. 
\subsubsection{Design} 

\begin{figure*}
    \centering
     \includegraphics[width=\textwidth]{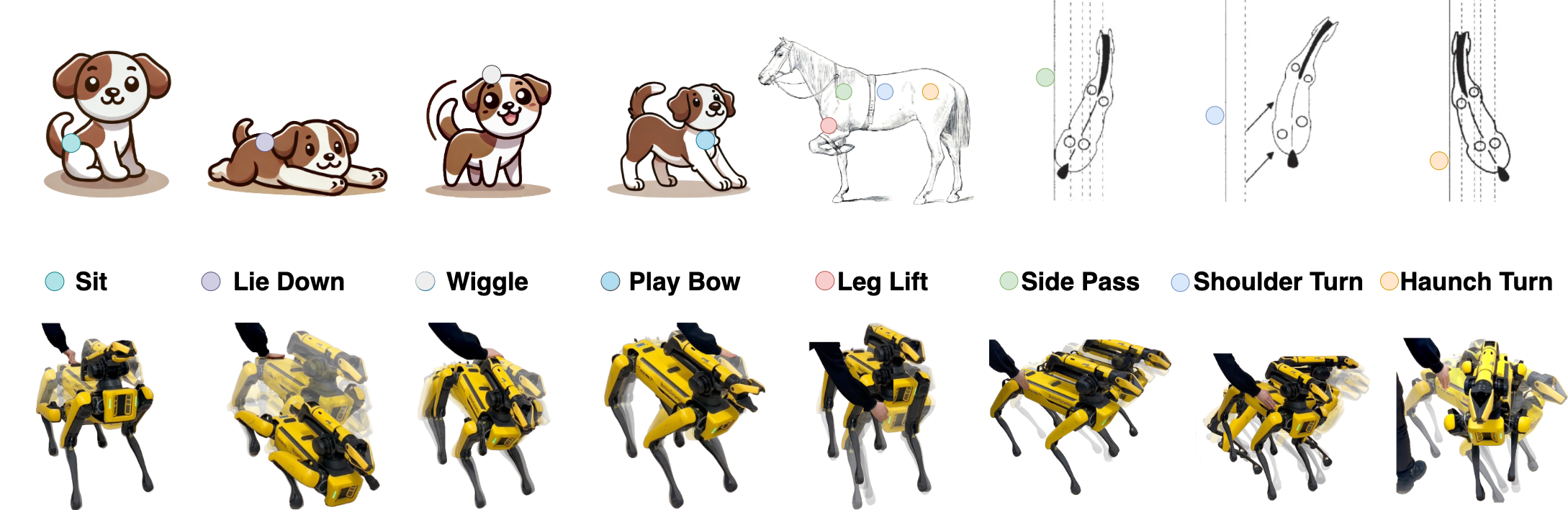}
    \caption{
    \textbf{pHRI deployment on Spot.}
    The first row illustrates the inspiration from human-animal interactions, showcasing how dogs and horses respond to touch cues (colored dots). 
    The second row depicts the corresponding robotic responses in deployment on Spot. 
    Video illustrations can be found in our supplementary material.
    }
    \label{fig:spot_hri_demo}
\end{figure*}

\begin{figure}
    \centering
        \includegraphics[width=\linewidth]{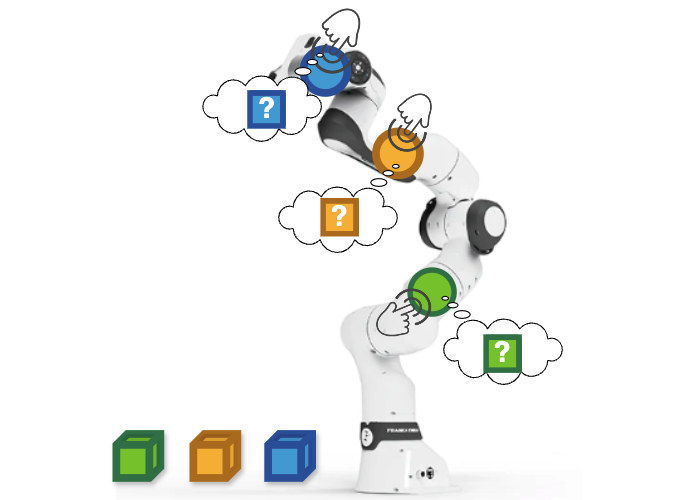}
    \caption{\textbf{Robot arm interactions design.} 
    With \shortname, we can assign ``virtual buttons'' for operators to provide manipulation instructions at no cost.
    } 
    \label{fig:franka_hri}
\end{figure}
\begin{figure}
    \centering
    \captionsetup{type=figure}
     \includegraphics[width=\linewidth]{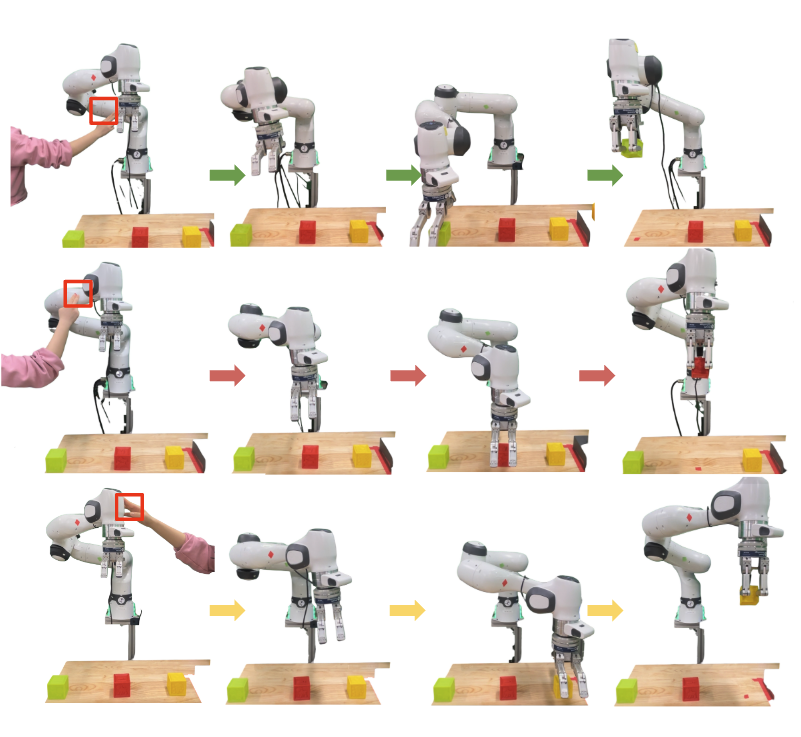}
    \caption{\textbf{pHRI deployment on Franka.} Each row illustrates Franka picking up a block of a different color.}
    \label{fig:franka_hri_deploy}
    \vspace{-2em}
\end{figure}
Using Spot, we program primitive actions using its Choreography SDK.
Based on the predictions from \networkname, we segment the robot’s body into distinct regions, each triggering a specific action (Fig.~\ref{fig:spot_hri}). 
We divide Spot actions into three categories: 1) Motion actions; 2) Posture change; and 3) Body expression. 
As a standardization of terminology, we refer to the five sides of its body as front, rear, left, right, and top. 
All motion actions are inspired by human-equine interactions~\cite{mcgreevy_chapter_2004}, while posture changes and body expression movements are derived from human-canine interactions.

\textbf{Motion Actions.} 
Motion actions involve whole-body movement and are based on horse training techniques that use physical cues to guide motion. These actions respond to touch on the side faces of the robot.
\begin{itemize}
    \item \textit{Turning on the Forehand} is triggered by touching the upper frontal section, prompting Spot to turn in the opposite direction by stepping with its front legs while its hind legs step in place. 
    \item \textit{Turning on the Haunches} occurs when the upper dorsal section is touched, making Spot step sideways while keeping its front legs stationary. 
    \item \textit{Shifting on Forehand/Haunches} results from touch on the lower frontal or dorsal sections, causing a weight shift in the opposite direction, mimicking a horse’s response to abdominal pressure. 
    \item \textit{Sidepassing} is activated by touch on the mesial section, prompting a lateral step while maintaining body alignment.
    \item \textit{Leg Lifting} occurs when the front or rear sections are touched, making Spot lift the corresponding leg, similar to a horse raising its hoof when handled by an equestrian.
\end{itemize}
    
\textbf{Posture Change.}    
We define posture change as Spot adjusting its body posture from a standing position, inspired by canine social signaling.
These responses are associated with touch on the top side of Spot’s body, excluding the front half, which is blocked when the arm is stowed. 
\begin{itemize}
    \item \textit{Lying Down} is triggered by touching the middle section, causing Spot to fully lower itself. 
    \item \textit{Sitting} occurs when the rear section near the hip is touched, prompting Spot to lower its hindquarters, similar to a dog sitting when patted on the hip.
\end{itemize}

\textbf{Body Expression.} 
Body expressions represent robot actions that replicate a canine’s expressive responses through movement. These behaviors are triggered by touch on Spot’s arm, which corresponds to the head and upper torso of a canine. 
\begin{itemize}
    \item \textit{Wiggle} occurs when the arm is touched, causing Spot to sway its body, similar to a puppy reacting to a pat on the neck.
    \item \textit{Play Bow} is triggered by touch near the gripper, making Spot sway while opening its gripper, mimicking a dog playfully bowing to welcome a friendly pat.
\end{itemize} 
    
\subsubsection{Spot Demonstration}

The quadruped choreography results show promising applications that are enabled by \shortname as shown in Fig.\ref{fig:spot_hri_demo}.\\

\subsection{Robotic arm}
\label{franka_phri}
Equipping robot arms with touch sensing capabilities provides significant benefits across multiple domains, improving safety, adaptability, and usability in human-robot interaction and manipulation tasks~\cite{argall_survey_2010}. 
Joint torque sensors are available in many robot arms. 
With touch sensing capabilities, one potential application is replacing physical buttons with virtual buttons to provide manipulation instructions. 

\subsubsection{Design}
Similar to seminal work~\cite{iskandar_intrinsic_2024}, we focus on human-robot interaction in manipulation tasks, where the robot selects and picks up one of three blocks based on touch input.

We randomly select three locations on the FR3 to attach green, red, and yellow stickers. 
The green sticker is attached at a lower joint link close to Franka's base. 
The red sticker is attached to its joint link at medium height, and the yellow sticker is attached close to the end effector.
We use these stickers to mimic ``buttons'' that will activate the Franka robot to pick up blocks of the same color from the desk in front of it (Fig.~\ref{fig:franka_hri}).

\subsubsection{Franka Demonstration}
We demonstrate the Franka tactile interactions in Fig.~\ref{fig:franka_hri_deploy}. 
Each row corresponds to pressing a different "button", where Franka is prompted to pick up the green, red, and yellow blocks, respectively.

\section{Conclusion}
We present \shortname, a whole-robot touch sensing method that uses only built-in joint sensors to localize contact in real time. 
Evaluated on both Franka arm and Spot quadruped, our data-driven approach achieves average localization errors of about 7–8 cm at 2,000 Hz, generalizing well across different robot instances. 
Furthermore, our pHRI demonstrations - such as quadruped choreography and robotic arm manipulation via virtual touch buttons - highlight the practical benefits of our approach for natural human-robot interactions.
\shortname offers a robust, easy-to-deploy alternative to dedicated tactile hardware, paving the way for more natural human-robot interactions.
In our future work, we aim to scale up data collection to support more robust and multi-contact predictions.

\section{Acknowledgments}
This work is supported by the Office of Naval Research (ONR) grant \#N00014-22-1-259. 
Wanjia is supported by Brown Undergraduate Teaching and Research Awards (UTRA) and Randy Pausch Fellowship.

\bibliographystyle{plainnat_custom}
\bibliography{custom, julia, zotero}

\end{document}